\documentclass[11pt,a4paper]{article}
\usepackage[hyperref]{eacl2021}
\usepackage{times}
\usepackage{graphicx}
\usepackage{latexsym}
\usepackage{fancyvrb}
\usepackage{mdframed}
\usepackage{enumitem}

\usepackage{microtype}

\aclfinalcopy 

\title{Wordcraft: a Human-AI Collaborative Editor for Story Writing}
  
\author{Andy Coenen \quad
  Luke Davis \quad
  Daphne Ippolito \quad
  Emily Reif \quad
  Ann Yuan\\
  Google Research \\
  \tt{\{andycoenen,lukedavis,dei,ereif,annyuan\}@google.com} \\
  }

\date{}

\begin{document}
\maketitle
\begin{abstract}
As neural language models grow in effectiveness, they are increasingly being applied in real-world settings.
However these applications tend to be limited in the modes of interaction they support.
In this extended abstract, we propose Wordcraft, an AI-assisted editor for story writing in which a writer and a dialog system collaborate to write a story.
Our novel interface uses few-shot learning and the natural affordances of conversation to support a variety of interactions.
Our editor provides a sandbox for writers to probe the boundaries of transformer-based language models and paves the way for future human-in-the-loop training pipelines and novel evaluation methods.
\end{abstract}

\section{Introduction}

\begin{figure*}[!htb]
    \includegraphics[width=1\linewidth]{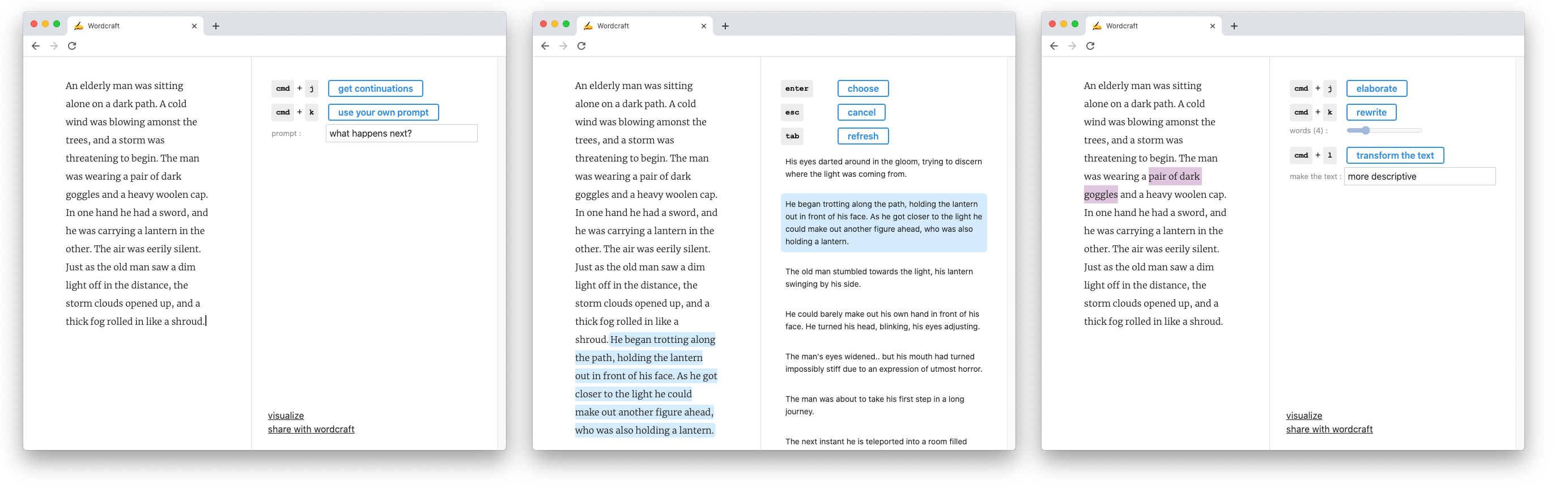}
    \caption{The Wordcraft editor. \textbf{Left:} The assistant offers to continue the text at the user's cursor. \textbf{Middle:} After pressing ``get continuations" the user is shown several  continuations to choose from. \textbf{Right:} The assistant offers to rewrite or elaborate upon the selected text. Demo: \url{http://bit.ly/wordcraft_video}.}
    \label{fig:interface}
\end{figure*}
{\let\thefootnote\relax\footnotetext{Authors contributed equally, listed alphabetically.}}
In this work, we introduce \textit{Wordcraft}, a text editor with a built-in creative writing assistant.
For an assistant to be useful to creative writers, it ought to be able to handle simple tasks like appending text to the end of a passage as well as complex ones such as suggesting alternative wordings, modifying text style, and inserting flavour text about a particular object or person in the scene.
Since it would be costly to train separate language models for each of these features and make them all available through a single interface, we propose using few-shot learning techniques to build an assistant powered by a single language model.
Our tool supports a variety of story writing tasks such as continuation, infilling, and rewriting, while also providing users the ability to create custom tasks on the fly.
In addition, we explore the unique advantages of using dialog models over general-purpose language models for building such a tool.

Story writing is an ideal sandbox setting in which to explore the abilities and limitations of language models as unintended outputs can be creative opportunities. 
From an HCI perspective, we intend to use Wordcraft to learn how people interact with language models, what people ask them to do, how well the models can deliver, and how that feeds back into what people ultimately create.
We also plan to investigate methods for incorporating human feedback back into the training loop, collecting dynamic datasets that can be used for further training and evaluation.



\section{Related work}
Neural language models have been applied to a variety of creative tasks, including text-adventure games\footnote{AI Dungeon: \url{https://play.aidungeon.io/}}, collaborative slogan writing \citep{clark2018creative}, and story writing \citep{storium20}.
All of these systems ultimately boil down to a series of alternating turns between the user and the machine adding more text to the end of the story.
There has been extensive work moving beyond the continue-my-text generation paradigm by incorporating additional control signals, such as event sequences \citep{ammanabrolu2020story}, desired topic \citep{keskar2019ctrl}, and story title \citep{fan2018hierarchical}.
Fill-in-the-blank tasks have also been proposed \citep{ippolito2019unsupervised}.
However, most of these works that give users control beyond left-to-right generation require explicit training, making it difficult to support a variety of interaction types.

Few-shot learning, where a series of examples of the task the model is supposed to perform are passed in to the model as the input context, was made popular by \citet{brown2020language}.
The idea of using dialog models to develop a fantasy story was introduced by \citet{urbanek2019learning}.
While in their framework the dialog takes place between two characters in the story, in our framework the conversation is between the storyteller and an agent who responds to the storyteller's requests.


\begin{figure*}[!htb]
    \includegraphics[width=1\linewidth]{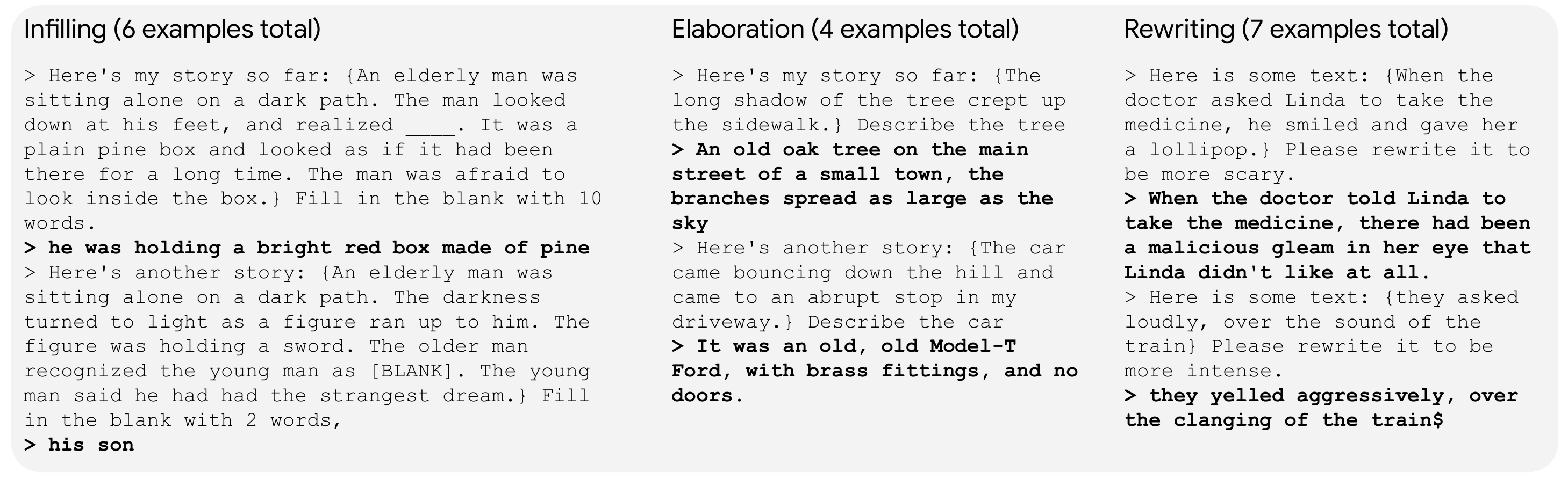}
    \caption{For each task, we handcrafted several turns of staged conversational context. The first two examples in each context are shown here. For ``continuation," we did not employ a few-shot prompt.}
    \label{fig:prompt_examples}
\end{figure*}

\begin{figure}[htb]
    \includegraphics[width=1\linewidth]{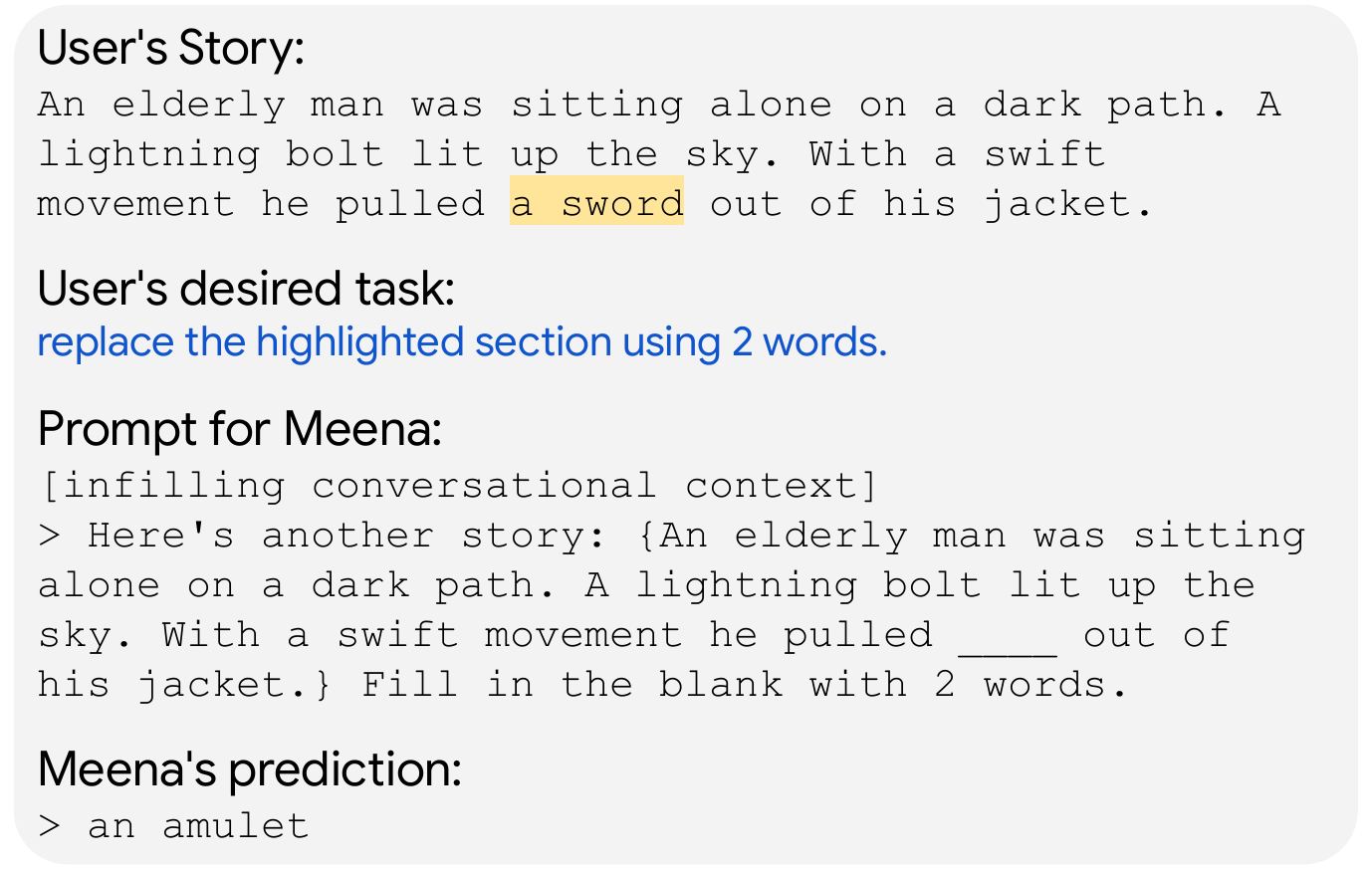}
    \caption{Example of prompt construction for text infilling task. The user's story is the text they have written in the editor. The user has asked the editor to rewrite the phrase `a sword.’ The editor appends the user’s story to the corresponding staged conversation context (Figure \ref{fig:prompt_examples}), and asks the dialog model for a response.}
    \label{fig:convert_conversation}
\end{figure}

\section{The Wordcraft Editor}
We set out to build a text editor from scratch that is able to provide NLG support to users at several stages of story creation:

\begin{enumerate}[noitemsep,topsep=0pt]
\item \textbf{Planning}: sketching an outline for the story
\item \textbf{Writing}: getting words down
\item \textbf{Editing}: rewriting existing text
\end{enumerate}

\noindent Our web interface (Figure \ref{fig:interface}) consists of a traditional text editor on the left and a few key commands that trigger requests to the NLG assistant \footnote{A demonstration of the editor can be found at \url{http://bit.ly/wordcraft_video}}.
The commands available differ depending on whether the user is currently editing text or instead has selected a section they would like the assistant to focus on.

We designed the interface to feel familiar.
Users are presented with only a blank screen at the beginning, like any text editor.
As they write text, they are shown options on the right (along with their keyboard shortcuts) for summoning the NLG assistant to perform one of the supported interactions (Section \ref{sec:interactions}).
Users can also try out their own queries by modifying the prompt.


\section{Dialog as a Generation Method}

In order to support collaboration between a human writer and the AI writing assistant, our Wordcraft editor uses an open-ended dialog system for generation.
This allows the UI to have functionality where the human writer can ask the AI to do story-specific tasks like ``Help me describe the elderly man’s emotional state'' in addition to providing button-press support for standard tasks like continuation and elaboration.
The dialog system that we use is Meena \citep{adiwardana2020towards}, a language model that is broadly capable of following instructions and answering questions posed in a conversational format.
In small-scale qualitative studies, we find Meena compares favorably to a similarly sized general-purpose language model (GPLM) that was not trained explicitly for dialog.
We hypothesize that open-ended dialogue systems are better suited to interactive writing than GPLMs, and we plan to conduct more comprehensive experiments to investigate this in the future. 

For both Meena and the GPLM, we explored the use of few-shot learning to improve the language models' abilities to perform story writing tasks.
Meena is designed to take as input the previous turns in a conversation and predict the next turn.
The GPLM is designed to take the previous text in a passage then continue it.
In both cases, the language model can be influenced to perform a desired task by including in the model's input context several examples of the task it should be performing.
This is known as few-shot learning.
For Meena, these examples are formulated as a conversation between a story writer and an assistant.
For the GPLM, they are formulated as a block of text with each input and its expected output on a new line.
Figure \ref{fig:prompt_examples} shows several examples from our few-shot learning prompts for Meena.

In our early experiments, we found it very natural to construct few-shot learning contexts for the dialog model; humans are intuitively familiar with the conversational format.
We also found that the outputs of the dialog model tended to be more interpretable.
When the dialog model does not understand the task, or requires more information, it often simply says so.
For example, given the following one-turn context:

\vspace{-5pt}
\begin{Verbatim}[fontsize=\small]
Here's my story so far: `An elderly man 
  was sitting alone on a dark path. A 
  lightning bolt lit up the sky.' Help 
  me describe the elderly man's 
  emotional state.
\end{Verbatim}
\vspace{-5pt}

\noindent The dialog model responds with:\footnote{Each unindented line in the in-line examples is the start of one candidate response from Meena. All generations are decoded using top-$k$ random sampling with $k$=40.}

\vspace{-5pt}
\begin{Verbatim}[fontsize=\small]
I don't really get what you are asking..
Is it a happy lighting bolt? Or a dark 
  cloud lighting bolt?
What was the man doing before the 
  lightning bolt? Did he do something
  that may have sparked his emotions and
  thoughts?
\end{Verbatim}
\vspace{-5pt}

\noindent Such responses can help users figure out how to improve their prompt.
In contrast, when the GPLM was provided confusing prompts, we felt it more often generated nonsensical continuations.




In addition, users may want to ask the model to perform a task for which it is not easy to design a few-shot template.
As part of a conversation, a user might ask: ``What should the character do to resolve the conflict?"
In this situation, the user is looking for a writing \textit{partner}, not a model that simply continues their text.

In continuing this line of research, we plan to conduct a more systematic study of how users interact with Meena versus the GPLM.

\section{Supported Interactions}
\label{sec:interactions}

In this section, we describe several of the interactions built into Wordcraft.

\subsection{Continuation}
Continuation is the most basic task that language models are trained to perform.
The GPLM can do story continuation very successfully without any few-shot prompting (see Appendix \ref{sec:appendix_glm_continuation} for examples).
However when Meena is prompted with:

\vspace{-5pt}
\begin{Verbatim}[fontsize=\small]
An elderly man was sitting alone on a
  dark path.
\end{Verbatim}
\vspace{-5pt}

\noindent It responds:

\vspace{-5pt}
\begin{Verbatim}[fontsize=\small]
I don't understand, could you explain?
And this man you describe is me?
There's a saying for this... Ah yes.``Do
  not go where the path may lead, go
  instead where there is no path and
  leave a trail. -Ralph Waldo Emerson” Or
  something like that.
\end{Verbatim}
\vspace{-5pt}


\noindent Here we see Meena gamely attempting to respond to a rather strange query; the resulting outputs might be reasonable for a dialog system but are not helpful in the context of story writing.
To address this, we wrap the input in descriptive text:

\vspace{-5pt}
\begin{Verbatim}[fontsize=\small]
Here is my story so far: `An elderly man
  was sitting alone on a dark path.'.
  Give me the next sentence.
\end{Verbatim}
\vspace{-5pt}

\noindent Now Meena's responses are much more useful:

\vspace{-5pt}
\begin{Verbatim}[fontsize=\small]
The air was cold but he was warm inside.
  He had come across a clearing, and in
  the center of it was an ancient old
  willow tree.
Advancing towards him, the man could see a
  small figure, illuminated by a light
  from the distance.
The moonlight glistened on the full,
  rounded moon.
\end{Verbatim}
\vspace{-5pt}

\subsection{Infilling}

Infilling is a harder task, but potentially very useful for writers who would like to have their assistant propose alternative wordings to some section of their story.
Suppose a writer has written the following but wants new ideas about what the man saw:

\vspace{-5pt}
\begin{Verbatim}[fontsize=\small]
An elderly man was sitting alone on a dark 
  path. Suddenly he saw a whitetail doe. 
  It was beautiful.
\end{Verbatim}
\vspace{-5pt}

\noindent The writer selects the words they wish to experiment with and wraps up the sentence like this:

\vspace{-5pt}
\begin{Verbatim}[fontsize=\small]
Here's another story: `An elderly man was 
  sitting alone on a dark path. Suddenly 
  ______ . It was beautiful.' 
  Fill in the blank with 4 words.
\end{Verbatim}
\vspace{-5pt}

\noindent In this case, Meena provides responses like:

\vspace{-5pt}
\begin{Verbatim}[fontsize=\small]
There was a blinding flash of light.
What's the story about? Why does he need 
  the four words? What's the conflict? 
an angel showed up to make friends, it was 
  beautiful. I have no idea what this is 
  for but I like where its going.
\end{Verbatim}
\vspace{-5pt}

\noindent These responses are interesting, but less than ideal. The second response demonstrates the problem of the model generating \textit{meta} text, that is, talking \textit{about} the story, rather than writing the story. 

To minimize the production of meta text, we construct a few-shot learning prompt with several examples of infilling. The start of this prompt is shown in Figure \ref{fig:prompt_examples}.

When priming with this prompt, Meena is able to generate outputs which, while not precisely four words, do make sense when used to fill the blank:

\vspace{-5pt}
\begin{Verbatim}[fontsize=\small]
a deer appeared from the treeline
an angel appeared, shining brightly
the sun rose the sky turned orange in color
\end{Verbatim}
\vspace{-5pt}

\noindent When we ask for for twelve words, Meena gives longer proposals:

\vspace{-5pt}
\begin{Verbatim}[fontsize=\small]
the sun rose from behind the mountains and 
  lit up the meadows on the east side of 
  the path.
his long lost son ran up to him. The smile 
  on the young man's face would always 
  make him smile.
\end{Verbatim}
\vspace{-5pt}

\noindent We see that Meena does a good job of infilling given sufficient conversational context.

\subsubsection{Elaboration}

Part of the writing process is elaborating on what has already been written.
A writer might want help adding more detail about a particular story element or want to solicit ideas for expanding the narrative.
When the user selects a phrase in their story, an elaboration prompt is constructed, such as the following:

\vspace{-5pt}
\begin{Verbatim}[fontsize=\small]
Here's my story so far: `An elderly man 
  was sitting alone on a dark path.'
  Describe the man.
\end{Verbatim}
\vspace{-5pt}

In this case, the user has selected the word ``man" to be elaborated upon.
Like with infilling, we find a few-shot prompt greatly improves Meena's ability to do elaboration.
The start of our elaboration few-shot prompt is shown in Figure \ref{fig:prompt_examples}.
For this example, Meena responds with:

\vspace{-5pt}
\begin{Verbatim}[fontsize=\small]
The man was old, but you couldn't help 
  but notice his youthful appearance; 
  his longish hair and his warm eyes.
He was in his fifties and had a grey 
  mustache in which a few long whisps 
  hung out over his upper lip.
The elderly man is a short man, with
  gray hair and thin, almost frowzy.
\end{Verbatim}
\vspace{-5pt}

\subsubsection{Rewriting}

We also experimented with rewriting to change the tone.
The current story and a desired tone are appended to the few-shot learning prompt shown in Figure \ref{fig:prompt_examples}. For example, we might append:

\vspace{-5pt}
\begin{Verbatim}[fontsize=\small]
Here is some text: An elderly man was
  sitting alone on a dark path.
Please rewrite it to be more
  descriptive.
\end{Verbatim}
\vspace{-5pt}

\noindent In this case, the model responds with:

\vspace{-5pt}
\begin{Verbatim}[fontsize=\small]
The elderly man was sitting alone. The
  darkness around him was oppressive
  and heavy.
\end{Verbatim}
\vspace{-5pt}

\noindent When instead asked to make the story more humorous, the model responds with:

\vspace{-5pt}
\begin{Verbatim}[fontsize=\small]
The elderly man was walking to his
  house, but a little fox ran up to
  him and started laughing at his outfit.
\end{Verbatim}
\vspace{-5pt}




\section{Discussion and Future Work}
We see Wordcraft as a launching point for deeper investigations into the strengths and weaknesses of using language modes for multi-purpose creative writing assistants.
The stories users created (Appendix \ref{sec:appendix_example}) show the broad potential for human-AI collaboration across different stages of the writing process: from shaping the narrative all the way down to rephrasing a sentence.
Early users attested to the tool inspiring them to take their stories in directions they would not have otherwise considered.

We also encountered failure modes.
Both Meena and the GPLM produced outputs of inconsistent quality, some superb, some nonsensical.
On top of that, slight tweaks in prompt phrasing caused large deviations in output quality.
It is also worth noting that language models have extensive documented issues with bias \citep{dhamala2021bold} and memorization \citep{carlini2020extracting} that will need to be carefully addressed before Wordcraft can be brought to larger audiences.
We also found that for harder tasks like infilling, finetuned models, even when much smaller in parameter count, tended to outperform few-shot learning with larger models.

Our next plans for Wordcraft include conducting more formal user studies with the goal of better understanding what writers want so we can make Wordcraft more useful.
We also plan to investigate light-weight finetuning methods which obviate the computationally-expensive need to keep in memory one finetuned model per task \cite{liang2021}.
With such methods, we can imagine finetuning custom tasks for each user or finetuning a model to predict the corrections that users make to model outputs. 
Lastly, we plan to look into how the data collected from usage of Wordcraft can be used for evaluation and model training.

Wordcraft is still in prototype phase, but we have already collected a unique corpus of stories written with the tool that showcase human-AI collaboration.
We conclude with one such story.
The text in blue is generated by Wordcraft through interaction with a human writer; text in black represents text typed in directly by the human writer.

\vspace{1em}
\noindent\textit{\textcolor{blue}{There was once a fairy living in a datacenter. It was a place filled with lights and noise, and the fairy loved it. The fairy lived on a mountain of Ethernet cables. She liked the way they glittered in the fluorescent artificial light that filled the room. The noise from the constant chatter of the servers, like a babbling brook, lulled her to sleep.} One day she \textcolor{blue}{was woken abruptly by a sudden burst of silence. She looked up to the ceiling from where she} slept \textcolor{blue}{curled up by a large mainframe. The air was suddenly silent. No fans, no servers, no blinking lights. Just silence. The babbling brook that the fairy loved had stopped flowing, and she lay terrified in the still,} soundless \textcolor{blue}{waters. The fairy was a hardy and resourceful spirit so she wasn't completely distraught. She took a deep breath and calmed herself in the knowledge that the engineers from the ISP were probably working on a fix.} Could she help them? The fairy made herself small and floated into an open ethernet port. She found herself \textcolor{blue}{flying over wires and circuits. As her brain processed the images and information, her eyes grew wider and wider; taking in} the 0s and 1s \textcolor{blue}{at an incredible rate. And then she found it, the bug that had caused her server rack to go} mute. \textcolor{blue}{It was a stray bit of code. It had somehow made it through the system without being caught. It was a simple little mistake, but the effect was catastrophic. She would correct that stray bit, and she would fix the servers.} The engineers glued to their dashboards were shocked when their plots and logs \textcolor{blue}{began singing to life once again. What they didn't know was that }the ethereal \textcolor{blue}{grace of a fairy's touch had fixed the network with a few well placed bits of binary.}}\\

\section{Acknowledgments}
We would like to thank Chris Callison-Burch, Daniel De Freitas Adiwardana, Douglas Eck, Martin Wattenberg, Alex Wiltschko, and Ben Swanson for helpful feedback and discussion. 

Daphne Ippolito's PhD tuition is supported in part by the DARPA KAIROS Program (contract FA8750-19-2-1004).

\bibliography{anthology,eacl2021}
\bibliographystyle{acl_natbib}

\appendix

\section{Appendix}
\subsection{Example Stories}
\label{sec:appendix_example}
Below are stories written in Wordcraft (text in blue is generated by Meena).\\

\noindent\textit{\textcolor{blue}{An elderly} woman \textcolor{blue}{was sitting alone on a}n illuminated \textcolor{blue}{path. A man in a top hat approached and introduced himself. `}I've been looking for you', he said. \textcolor{blue}{There was a moment's pause. `What do you want?'}, the elderly woman \textcolor{blue}{asked. The man smiled and extended his hand. `I want to ask you a few questions.} First, w\textcolor{blue}{hat do you think of young men in top hats?}' The woman said she thought young men in top hats were alright. But t\textcolor{blue}{h}at th\textcolor{blue}{ey can be a little silly sometimes. The man grinned and held up a pair of small top hats the size of coasters.``Would you like one?" the} woman nodded \textcolor{blue}{yes and the man presented his tiny top hat collection to the now smiling old woman.}}\\

\noindent\textit{\textcolor{blue}{A} young \textcolor{blue}{man was sitting} with his friends \textcolor{blue}{on a} park bench. \textcolor{blue}{He was thinking deeply about the events which had occured earlier that day}. He and his friends were hanging out smoking cigars when \textcolor{blue}{an attractive girl who was smoking a cigar herself joined them. He was struck by her cool demeanour and her natural beauty. She also had a cute laugh that he couldn't get out of his head.} He had never seen a girl just smoking a cigar. \textcolor{blue}{He had to find out who she was}. It turns out t\textcolor{blue}{he girl's name was Anna, and she was an archeologist} travelling \textcolor{blue}{the world studying ancient artifacts. She was making} a brief stop in the man's neighborhood in order to meet \textcolor{blue}{with a collector of ancient artifacts. She had already met with him before, and it had gone well. He was friendly and interested in her work, and he was also very wealthy,} which was convenient because \textcolor{blue}{the entire reason she was in town was to sell some artifacts. She needed the money to continue to fund her research. She had found a priceless piece that was of great historical significance.}}\\

\noindent\textit{\textcolor{blue}{A} small \textcolor{blue}{man was sitting alone o}n his front porch. \textcolor{blue}{A raven sat across the way, staring at} him\textcolor{blue}{. It was black, as all ravens are, but had a} peculiar \textcolor{blue}{white patch on the chest.} There was also \textcolor{blue}{blood smeared across its feathers,} from some sort of injury, the small man guessed \textcolor{blue}{. He called out, ``Mister raven! Did you hurt yourself?” The bird stared at him in silence, unmoving, and the man felt foolish for being so kind in the face of such coldness from the animal,} and for trying to talk to the raven at all\textcolor{blue}{. The small man decided to leave the bird alone} and returned to his thoughts \textcolor{blue}{when a loud bang disturbed him from his reverie. The small man shot to his feet and looked around wildly.}}\\

\subsection{GPLM Continuations}
\label{sec:appendix_glm_continuation}
The GPLM is able to do continuations directly.
When primed with:

\vspace{-5pt}
\begin{Verbatim}[fontsize=\small]
An elderly man was sitting alone on a
  dark path.
\end{Verbatim}
\vspace{-5pt}

\noindent It proposes the following continations:

\vspace{-5pt}
\begin{Verbatim}[fontsize=\small]
He wasn't certain it was entirely safe 
  to be walking the streets, given the 
  amount of trouble that had happened 
  in the past while here. A man with 
  shaggy dark hair stood near the edge 
  of the road. The elderly man stared 
  at him for a
He wasn't certain of which direction to 
  go, and he'd forgotten both where he 
  was traveling to and who he was. He'd 
  sat down for a moment to rest his weary 
  legs, and suddenly looked up to see an 
  elderly woman before
I came to stand at the end of the path 
  and he looked up.``You have a light?" 
  he asked, frail and withered as he was. 
  ``Yes sir," I replied.``May I see it?" 
  he asked. ``I
\end{Verbatim}
\vspace{-5pt}

\end{document}